%% file: l4dc2023-sample.tex
\documentclass[12pt, final]{l4dc2023}

\usepackage{amsfonts}
\usepackage{bm}
\usepackage{graphicx}
\usepackage{mathtools}
\usepackage{hhline}
\usepackage{multirow}
\usepackage[font=small]{caption}
\newcommand{\STAB}[1]{\begin{tabular}{@{}c@{}}#1\end{tabular}}

\usepackage{epstopdf}
\AppendGraphicsExtensions{.tiff}

\DeclareMathOperator{\sech}{sech}
\newcommand{\norm}[1]{\left\lVert#1\right\rVert}

\title[Fourier Koopman Operator]{Online Estimation of the Koopman Operator Using Fourier Features}
\usepackage{times}
\author{%
 \Name{Tahiya Salam} \Email{tsalam@seas.upenn.edu} \\
 \Name{Alice Kate Li} \Email{alicekl@seas.upenn.edu} \\
 \Name{M. Ani Hsieh} \Email{mya@seas.upenn.edu} \\
 \addr{GRASP Lab, University of Pennsylvania, Philadelphia, PA}
}

\begin{document}

\maketitle

\begin{abstract}%
Transfer operators offer linear representations and global, physically meaningful features of nonlinear dynamical systems. Discovering transfer operators, such as the Koopman operator, require careful crafted dictionaries of observables, acting on states of the dynamical system. This is ad hoc and requires the full dataset for evaluation. In this paper, we offer an optimization scheme to allow joint learning of the observables and Koopman operator with online data. Our results show we are able to reconstruct the evolution and represent the global features of complex dynamical systems.
\end{abstract}

\begin{keywords}%
  Koopman Operators, Learning Nonlinear Dynamics, Nonlinear Identification%
\end{keywords}

\section{Introduction}
\label{sec:intro}
\input{sections/introduction}
%===============================================================================

\section{Related Works}
\label{sec:related works}
\input{sections/related-works}
%===============================================================================

\section{Preliminaries}
\label{sec:preliminaries}
\input{sections/preliminaries}
%===============================================================================

\section{Methodology}
\label{sec:methodology}
\input{sections/estimating-koopman}

\section{Experimental Results}
\label{sec:result}
\input{sections/experimental-results}
%===============================================================================

% \section{Miscellaneous}
% \label{sec:misc}
% \input{sections/misc}

%===============================================================================

\section{Conclusion}
\label{sec:conclusion}
\input{sections/conclusion.tex}

\section{Acknowledgements}
\label{sec:acknowledgements}
\input{sections/acknowledgements.tex}

\bibliography{references} 

\end{document}

%% file: sections/introduction.tex
In dynamical systems theory, a fundamental goal is data-driven state prediction and feature representation. Given collected sensor measurements over time, we often want to characterize the current nonlinear system states and predict the future states. Furthermore, for applications such as robotics, sensing data is collected in real-time and contains noise. In this work, we will focus on online methods for extracting global dynamics of the underlying system with the Koopman operator. 

Many state estimation techniques for systems whose dynamics are governed by differential equations rely on extracting dynamics from data. A powerful lens for estimating system dynamics and their representations is by characterizing the global dynamics of an ensemble of sensor observations or measurements \citep{Mezic2005, Klus2020EigendecompositionsSpaces}. A useful mathematical tool for understanding the time evolution of states and their global dynamics is transfer operator theory, through which we can study the action of a dynamical system on mass densities of initial conditions. The transfer operator is defined on some infinite dimensional linear functional space and describes a linear time evolution of the transformed densities. This is desirable as the original, nonlinear dynamics are lifted to a space such that the transformed densities evolve linearly. For example, the global feature tracking of trajectories in flow-like environments, analyzing pair-wise relationships in text documents, or observation of conformational changes in metastable walking may be suitable for study under transfer operator theory \citep{Salam2022LearningAwareness, Klus2020EigendecompositionsSpaces, Costa2021MaximallyData}. There are many methods for extracting the Koopman operator, a transfer operator in which the lifted densities are referred to as observables, from data \citep{Schmid2010DynamicData, Li2017ExtendedOperator, Klus2020EigendecompositionsSpaces}. Many of these data-driven Koopman construction techniques have been used with real sensing data \citep{Korda2018OptimalControl, Abraham2019ActiveOperators, Folkestad2020EpisodicLanding, Bruder2019ModelingControl, Salam2022LearningAwareness}. While these works are promising in their application of the Koopman operator to different domains, they are limited in that they derive the operator as a closed-form, rigid solution and require data across the entire time horizon.

% Representations for extracting dynamics inherently consider the time dependence of the states. Kalman filters, and its extensions, require knowledge of the system properties and measurement modality to produce state estimates \citep{Kalman1960AProblems, Jain2004AdaptiveNetworks, Cannell2005AVehicles}. Dimensionality reduction and spectral analysis techniques such as Proper Orthogonal Decomposition (POD) and Dynamic Mode Decomposition (DMD) \citep{Sirovich1987TurbulenceStructures, Schmid2010DynamicData} have been successfully applied as techniques that are still able to deduce properties of the dynamics without explicitly modeling the system and sensors \citep{Salam2019, Manohar2018Data-DrivenPatterns}. Gaussian processes (GPs) and neural networks also both allow for state representation without explicitly specifying the dynamics or sensor models \citep{Williams2006GaussianLearning, Schmidhuber2015DeepOverview}. However, GPs require a specification of the underlying dependence structure of the data to successfully reason about the evolving dynamics in real-time \citep{Xu2011AdaptiveNetworks, Krause2008Near-optimalStudies}. Neural networks have been widely used to represent the dynamics of the data and are particularly promising as they are universal function approximators \citep{Brunton2020MachineMechanics, Karniadakis2021Physics-informedLearning, Ong2016DynamicallyPM2.5}. 

This paper presents two fundamental advances in the estimation of the Koopman operator. First, we devise a way of jointly learning the observables and the Koopman operator from data collected by robots in a reproducing kernel Hilbert space using random Fourier features. Second, the joint learning is formulated such that construction of the observables and Koopman operator can be done completely online. We test proposed framework on dynamical systems with varying degrees of complexity, and show that it is desirable for the representation of complex dynamical systems; future state prediction; and computing physically meaningful eigenfunctions.

%% file: sections/related-works.tex
We review the literature related to the construction of the Koopman operator and emphasize some shortcomings of the existing methods.

\textbf{Closed-form solutions.} One of the most common data-driven methods for estimating the Koopman operator relies on Dynamic Mode Decomposition (DMD) \citep{Schmid2010DynamicData}. DMD is a spectral decomposition of the Koopman operator, for which the dynamics are assumed to be linear \citep{Mezic2005, Rowley2009SpectralFlows}. Extended Dynamic Mode Decomposition (EDMD) is a nonlinear generalization of DMD, where dynamics are assumed to be described using a nonlinear invertible transformation characterized by a dictionary acting on input data \citep{Williams2015ADecomposition}. In this work, instead of the Koopman operator acting directly on observables (in this case, the data from the system), the operator is acting on the dictionary applied to observables which results in an expanded set of observables. In EDMD, the user determines the choice of dictionary, tailored to the dataset.

Recent works have explored the connections between the Koopman operator and reproducing kernel Hilbert spaces (RKHS) \citep{Williams2015AAnalysis, Klus2020EigendecompositionsSpaces, Kawahara2016DynamicAnalysis, Das2018KoopmanSpaces}. Representations of the Koopman operator in RKHS allows for the analysis of the operator in any domain where there is a similarity measure given by a kernel. As the Koopman operator construction relies on an inner product computation, these methods allows for the computation of inner products implicitly through the use of a defined kernel function. Another advantage to these algorithms is that they give an approximation of the Koopman operator with a set of nonlinear basis functions due to the expressiveness of kernel functions.

\textbf{Dictionary learning approaches.} A more flexible approach formulates a trainable dictionary represented by an artificial neural network (NN) within the EDMD framework \citep{Li2017ExtendedOperator, Yeung2019LearningSystems, Mardt2020DeepConstraints}. Rather than the user defining a dictionary of observables, a dictionary can be learned using data from the system of interest. In many applications, it is not necessary to explicitly construct the Koopman operator but instead construct its eigenfunctions, a useful decomposition for understanding the stability of the process of interest \citep{Mauroy2016GlobalOperator}. There are many techniques focused on data-driven learning of Koopman eigenfunctions \citep{Korda2018OptimalControl, Folkestad2020EpisodicLanding, Folkestad2020ExtendedControl, Haseli2021LearningDecomposition, Kaiser2021Data-drivenControl, Sznaier2021AOperators, shi2021acd, leask2021modal}. 

In this work, we leverage insights from dictionary learning to automate the procedure for learning the transfer operators online. Here, we focus on learning the kernel transfer operators. Instead of black-box optimization to construct the dictionary of observables, we apply insights from the kernel methods literature to construct the dictionary. This method provides structure and meaning to the dictionary, but it is still an automated procedure similar to NN methods.

%% file: sections/preliminaries.tex
\begin{figure}
    \centering
\includegraphics[width=0.75\textwidth]{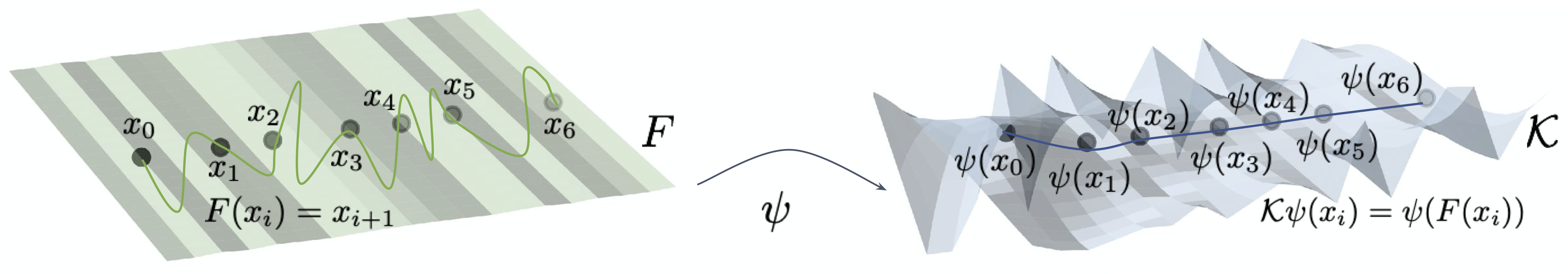}
    \caption{The state $\mathbf{x}_i$ evolves nonlinearly according to a dynamical system function $F$. Using the map $\psi$, the states can be lifted to an alternative space, where the lifted mappings of the states $\psi(\mathbf{x}_i)$ evolves linearly according to the Koopman operator $\mathcal{K}$.}
    \label{fig:lifting}
\end{figure}

Before proposing our strategy for learning the Koopman operator with Fourier features, we introduce the definition of the Koopman operator and connections between kernels and Fourier features.
\subsection{Koopman Operator}
The basic form of a discrete-time dynamical system in state-space form is 
\begin{equation} \label{eq:dynamical system}
    \mathbf{x}({t+1}) = f(\mathbf{x}({t})),
\end{equation}
where the function $f$ maps the state space in $\mathbb{R}^d$ to itself.  Given a class of scalar, complex valued functions $\mathcal{F}$, an observable is any $\psi(\cdot) : \mathbb{R}^d \rightarrow \mathbb{C}$, where $\psi$ maps the state of the system into a scalar. The value of the observable is determined by the state of the system and its evolution over time is described by the composition of the observable with the state dynamics $f(\cdot)$ as
\begin{equation} \label{eq:observable}
    \psi(\mathbf{x}({t+1})) = \psi(f(\mathbf{x}({t}))).
\end{equation}
The Koopman operator, $\mathcal{K}$, is applied to the observable and describe the evolution of observables, as in Eq. \eqref{eq:observable}, under the state evolution. As such, the evolution of observable $\psi$ can be written as $(\mathcal{K} \psi)(\mathbf{x}) = \psi(f(\mathbf{x}))$. The relationship between the dynamical systems representation and transfer operator can be shown in Fig. \ref{fig:lifting}. The observables can be thought of as a lifting operation that maps state dynamics into a different space, where observables can be propagated forward in time by $\mathcal{K}$. 

\subsection{Kernels}
Kernel functions $k: X \times X \rightarrow \mathbb{R}$ are applied to elements of some space, $X$, to measure the similarity between any pairs of elements. Let $\phi$ be the feature map associated with kernel $k$ defined on $X$. Then, define feature matrices as 
\begin{equation} \label{eq:feature-matrix}
    \Phi =  [\phi(x_1) \dots \phi(x_n)].
\end{equation}
The similarity metric in the form of a kernel function can be found by computing the inner product of features $\phi(x)$ acting on elements of the input space in some high-dimensional, possibly infinite, feature space in $\mathbb{R}^M$. Thus, kernels allow us to compare objects based on their features. Formally, for inputs $x, x' \in X$, a feature map $\phi: X \rightarrow \mathbb{R}^{M}$, and some valid inner product $\langle\cdot{ , }\cdot\rangle_\mathcal{V}$, a kernel $k$ is defined as $k(x, x') = \langle\,\phi(x), \phi(x')\rangle_\mathcal{V}$, where $\mathcal{V}$ is an inner product space.
Kernels on features spaces that are positive definite can be used to define a function $f$ on $X$. The space of such functions $f$, such that the evaluation of $f$ at $x$ can be represented as the inner product $\langle\,\phi(x), \phi(x')\rangle$ in feature space, is referred to as the Reproducing Kernel Hilbert Space \citep{Gretton2013IntroductionAlgorithms}.

% For an unknown kernel function $k$, let $\sigma_0$ be the kernel scale parameter, $\theta$ be a vector of all the hyperparameters of the model, including the kernel frequencies from their Fourier representation. Applying the feature maps to the data, we can define $\mathbf{\Phi} = [\phi(\mathbf{x}_1), \dots, \phi(\mathbf{x}_n)]$.

\subsection{Random Fourier Features}\label{subsec:RFFs}
Random Fourier Features (RFFs) are spectral domain representations of kernels \citep{Rahimi2009RandomMachines}. These techniques rely on representing the stationary covariance function as the Fourier transform of a positive finite measure, formalized by Bochner's Theorem and its corollary stated below for the sake of completeness.

\begin{theorem}\label{bochner-thm} (Bochner's Theorem) \citep{SalomonBochner1932VorlesungenIntegrale} Every positive definite function $\hat{\mu} : \mathbb{R}^{D} \rightarrow \mathbb{C}$ for all $\mathbb{\mathbf{x}} \in \mathbb{R}^D$ is the Fourier transform of a non-negative finite Borel measure $\mu$ on $\mathbb{R}^D$. That is, for any $\hat{\mu}(\mathbf{x})$ there exists a measure $\mu$ such that $\hat{\mu}(\mathbf{x}) = \int_{\mathbb{R}^D} e^{-i \mathbf{x}^\top \boldsymbol{\omega}} d\mu(\boldsymbol{\omega})$.
\end{theorem}

\begin{corollary} If $\hat{\mu}(0) = 1$ and the measure $\mu$ is a probability measure with probability density function (pdf) $f_{\Omega}$ on random variable $\Omega$ with realizations $\boldsymbol{\omega} \in \mathbb{R}^D$, then $\hat{\mu}(\mathbf{x} - \mathbf{x}') =: k(\mathbf{x}, \mathbf{x}')$ is a continuous stationary positive-definite covariance function $k(\mathbf{x}, \mathbf{x}') = \int_{\mathbb{R}^D} e^{-i (\mathbf{x} - \mathbf{x}')^\top \boldsymbol{\omega}} f_{\Omega}(\boldsymbol{\omega})d\boldsymbol{\omega}$.
\end{corollary}

While the dot product can be easily computed as the kernel evaluation, this means that the data can only be accessed through the evaluation of the kernel function or through a matrix containing all of the evaluations across all pairs of data points. However, this might be computationally expensive for large datasets. To alleviate this,  \cite{Rahimi2009RandomMachines} proposed explicitly mapping the data to a lower-dimensional Euclidean space using a randomized feature map $\hat{\phi}$. 
A known pdf can be represented using a finite number of Monte Carlo (MC) samples, where $\{\boldsymbol{\omega}_m\}_{m=1}^M \stackrel{i.i.d}{\sim} f_{\Omega}(\boldsymbol{\omega})$ lends to a finite dimensional approximation of the feature map $\hat{\phi}(\mathbf{x}) \in \mathbb{C}^{M}$ that is 
\begin{equation}
    k(\mathbf{x}, \mathbf{x}') \approx \frac{1}{M} \sum_{m=1}^{M} e^{-i(\mathbf{x} - \mathbf{x}')^\top \boldsymbol{\omega}_m} = \langle \hat{\phi}(\mathbf{x}), \hat{\phi}(\mathbf{x}') \rangle_{\mathbb{C}}.
\end{equation}
Then, the feature map can be decomposed such that $\hat{\phi}(\mathbf{x}) = {1}/{\sqrt{M}} [e^{-i\mathbf{x}^\top\boldsymbol{\omega}_1}, \dots, e^{-i\mathbf{x}^\top\boldsymbol{\omega}_M}] \in \mathbb{C}$. Since the kernels we will use are real valued, we can re-write the feature map to be $\hat{\phi}(\mathbf{x}) \in \mathbb{R}^M$
\begin{equation} \label{eq:feature-map}
    \hat{\phi}(\mathbf{x}) = \frac{\sqrt{2}}{\sqrt{M}}[\cos(\mathbf{x}^\top\boldsymbol{\omega}_1 + b_1), \dots, \cos(\mathbf{x}^\top\boldsymbol{\omega}_M + b_M)], \quad b_i {\, \sim \,} \text{Uniform}(0, 2\pi).
\end{equation}

%% file: sections/estimating-koopman.tex
\begin{figure}
    \centering
    \includegraphics[width=0.7\textwidth]{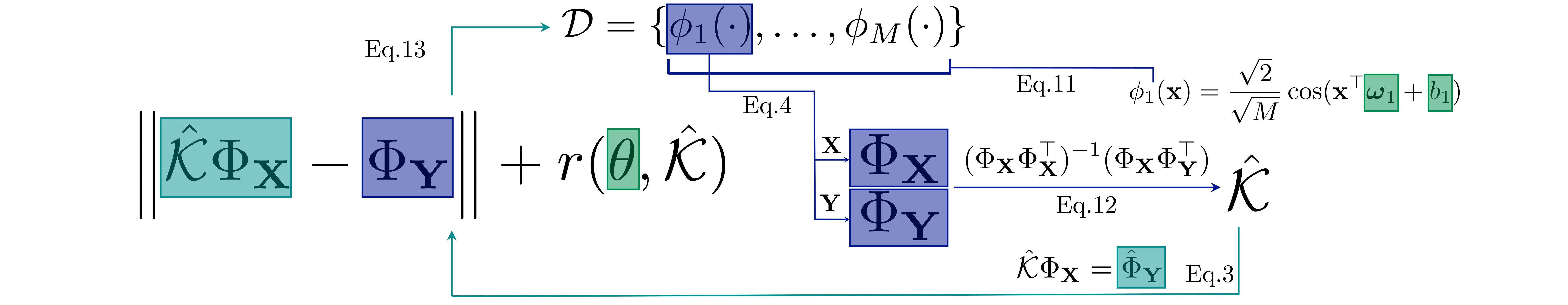}
    \caption{Overview of optimization framework for joint learning of dictionary of observables and Koopman operator. Random Fourier Features $\phi$ are initialized with weights $\omega, b$ and used to create dictionaries of observables $\Phi_\mathbf{X}$ and $\Phi_\mathbf{Y}$ on the training data $\{\mathbf{X}, \mathbf{Y}\}$. The Koopman operator is constructed using $\Phi_\mathbf{X}$ and $\Phi_\mathbf{Y}$ and estimates $\hat{\Phi}_\mathbf{Y}$. The loss is computed and the weights are updated in the dictionary. Dark blue values are calculated using the current weights in green, while light blue are the Koopman computed estimates.}
    \label{fig:loss-schematic}
\end{figure}

In this section, we present our main algorithmic contributions: the online joint learning of Koopman operator and observable dictionary. We first present a construction of the Koopman operator using kernels \citep{Klus2020EigendecompositionsSpaces} and demonstrate how we can bypass the need to explicitly specify a kernel by leveraging RFFs \citep{Rahimi2009RandomMachines}. In doing so, we can jointly learn the underlying Koopman operator and observable dictionary within an RKHS framework. 
%The structure provided by the RFFs allows us to characterize the observables and automate the process of construction a Koopman operator for a dynamical system without a neural network architecture.% % Took this part out because it seems to be mentioned more explicitly again but more explicitly at the end of this section: With the learned RFF feature maps, we can then construct the kernel Koopman operator explicitly using Eq. \eqref{eq:koopman-est}. % 
Using RFFs as the feature map on large datasets is more efficient than existing kernel EDMD methods, since RFFs involve mapping to a lower-dimensional space, instead of a higher dimension that kernels require. RFFs also provide the necessary kernel structure, with $(d+1)$ learnable parameters per RFF, and so our framework should require less training and tuning than an NN. Furthermore, new observations can sequentially be appended to training data, lending to an online learning schema. 

Let us re-write Eq. \eqref{eq:dynamical system} as $\mathbf{y}_i = f(\mathbf{x}_i)$. Suppose we have $N$ pairs of training data $(\mathbf{x}_i, \mathbf{y}_i)$ collected over the system of interest, where $\mathbf{x}, \mathbf{y} \in \mathbb{R}^d$. We herein refer to a data sample as a particle. Let $\mathbf{X} = [ \mathbf{x}_1,  \mathbf{x}_2, \dots \mathbf{x}_N ]^\top$ be a matrix of the $N$ particles at a specific time instance and $\mathbf{Y} = [ \mathbf{y}_1,  \mathbf{y}_2, \dots \mathbf{y}_N ]^\top$ be the position of the particles after time has elapsed. $\mathbf{Y}$ is then the time shifted version of $\mathbf{X}$. If the positions of the $N$ pairs of training data are tracked over time, we can index the $N$ particles at time $t$ by $\mathbf{X}(t)$ and the time shifted particles as $\mathbf{Y}(t)$. The training data can then be summarized as $\{ (\mathbf{X}(t), \mathbf{Y}(t)) \}_{t=1}^T$, where these can be thought of as the $T$-length trajectories of the particles.

Define feature maps $\Phi_{\mathbf{X}} = [ \phi(\mathbf{x}_1), \phi(\mathbf{x}_2), \dots,  \phi(\mathbf{x}_N)]^\top$ and $\Phi_{\mathbf{Y}} = [ \phi(\mathbf{y}_1), \phi(\mathbf{y}_2), \dots,  \phi(\mathbf{y}_N)]^\top$ on the particles using the feature maps in Eq. \eqref{eq:feature-matrix}. The feature map $\Phi_{\mathbf{Y}}$ can be thought of as a time shifted version of $\Phi_{\mathbf{X}}$. We can also construct the feature maps $\Phi_{\mathbf{X}(t)}$ and $\Phi_{\mathbf{Y}(t)}$ using the $N$ particles from a specific time instance. From \citep{Klus2020EigendecompositionsSpaces}, the Koopman operator associated with the dynamical system of interest can be estimated using the relationship 
\begin{equation}\label{eq:koopman-est}
    \hat{\mathcal{K}} = (\Phi_\mathbf{X}\Phi_\mathbf{X}^\top)^{-1}(\Phi_\mathbf{X}\Phi_\mathbf{Y}).
\end{equation}
This estimation of the Koopman operator leverages the kernel functional evaluation instead of the explicit feature map computation \citep{Klus2020EigendecompositionsSpaces, Williams2015AAnalysis}. These approaches assume that the dependence in the training data can be appropriately described by a kernel function, a computationally efficient operation. In the standard formulation of the RFFs, the distribution $f_{\Omega}(\boldsymbol{\omega})$ is known and corresponds to specific kernel forms. For example, sampling from a Gaussian distribution corresponds to the RBF kernel. 

In this work, instead of assuming a known form of the kernel function $k(\mathbf{x}, \mathbf{x}')$, such as the RBF, we instead approximate the kernel function using the RFFs, as in Eq. \ref{eq:feature-matrix},  described in Section \ref{subsec:RFFs}. Thus, instead of assuming some known form of the dependence within the data, we jointly the learn the structure of the feature maps and the corresponding Koopman operator. Let $\theta$ be the collection of weights $\boldsymbol{\omega}_{i}$ and $b_i$ associated with the RFFs that we would like to learn in order to determine the structure of the underlying feature maps that best describe the data. Previous work by \cite{Li2017ExtendedOperator} and \cite{Yeung2019LearningSystems} show this is achieved through the following minimization scheme
\begin{equation} \label{eq:min-loss}
    \min\limits_{\hat{\mathcal{K}}, \theta} \sum\limits_{t=1}^T \norm{\Phi_{\mathbf{Y}(t)} - \hat{\mathcal{K}} \Phi_{\mathbf{X}(t)}}_2 + \lambda_1 \norm{\hat{\mathcal{K}}}_2 + \lambda_2 \norm{\theta}_1.
\end{equation} 
We learn the dictionary of observables, where observables are expressed through the feature maps. This connection between dictionaries of observables and features maps was made when kernel EDMD was first proposed \citep{Williams2015AAnalysis}. In the work by \cite{Li2017ExtendedOperator}, the dictionary of observables is unknown and an NN is used to estimate them.  We use the kernel expression of the Koopman operator and leverage RFFs in capturing the dictionary of observables. The learned RFF feature maps and kernel Koopman operator are learnt via minimizing Eq. \eqref{eq:min-loss} via stochastic gradient descent and enforcing the kernel Koopman operator structure using Eq. \eqref{eq:koopman-est}. We note that any optimization scheme can be used. The loss schematic is summarized in Fig. \ref{fig:loss-schematic}.

%% file: sections/experimental-results.tex
We describe the details for simulating benchmark prototypical dynamical systems with varying degrees of complexity used to study transfer operators, our quantitative results evaluating the reconstruction of trajectories and eigenfunction approximation, and qualitative results depicting physically meaningful global features within these systems. 

\begin{figure}[!ht]
    \centering
    \subfigure{\includegraphics[width=0.32\textwidth]{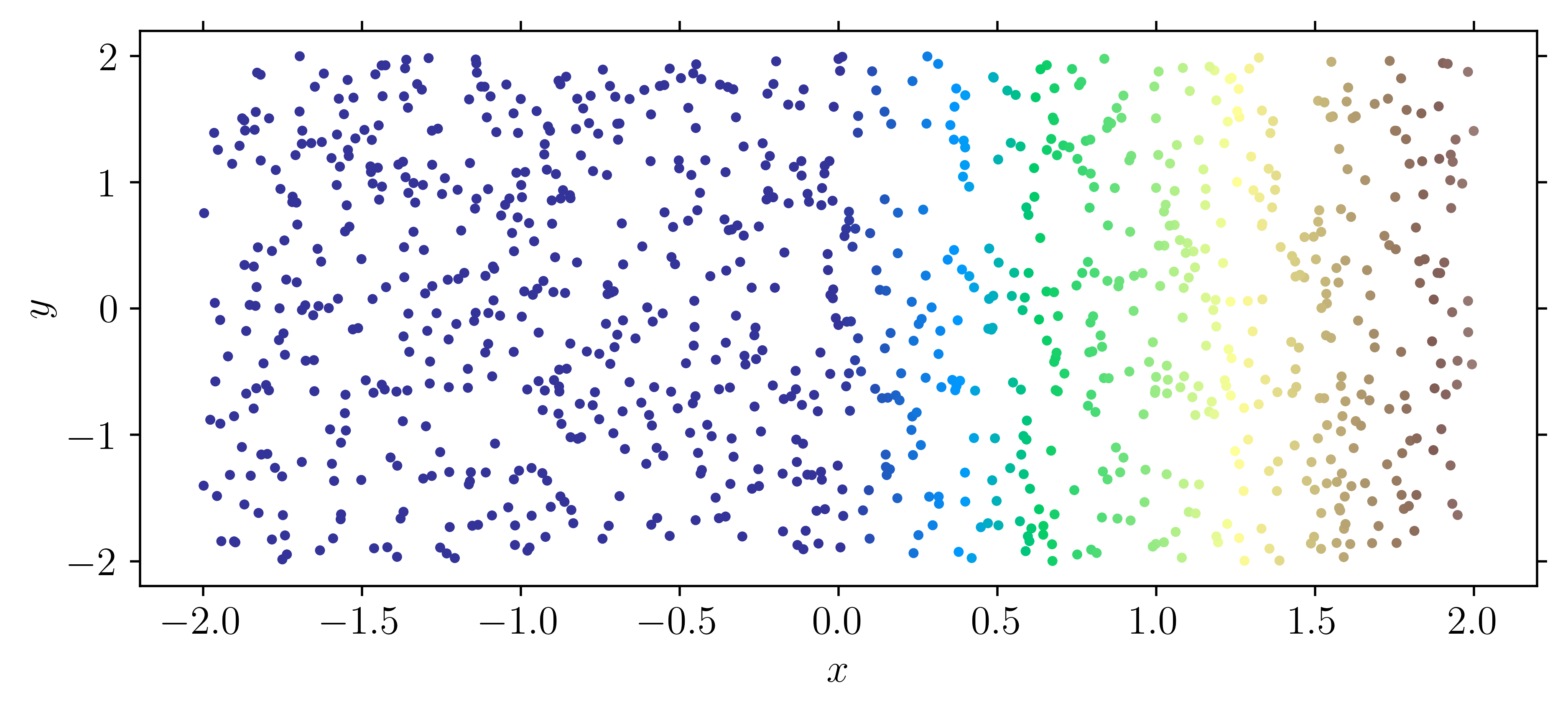}\label{fig:duffing-oscillator-init}} \subfigure{\includegraphics[width=0.32\textwidth]{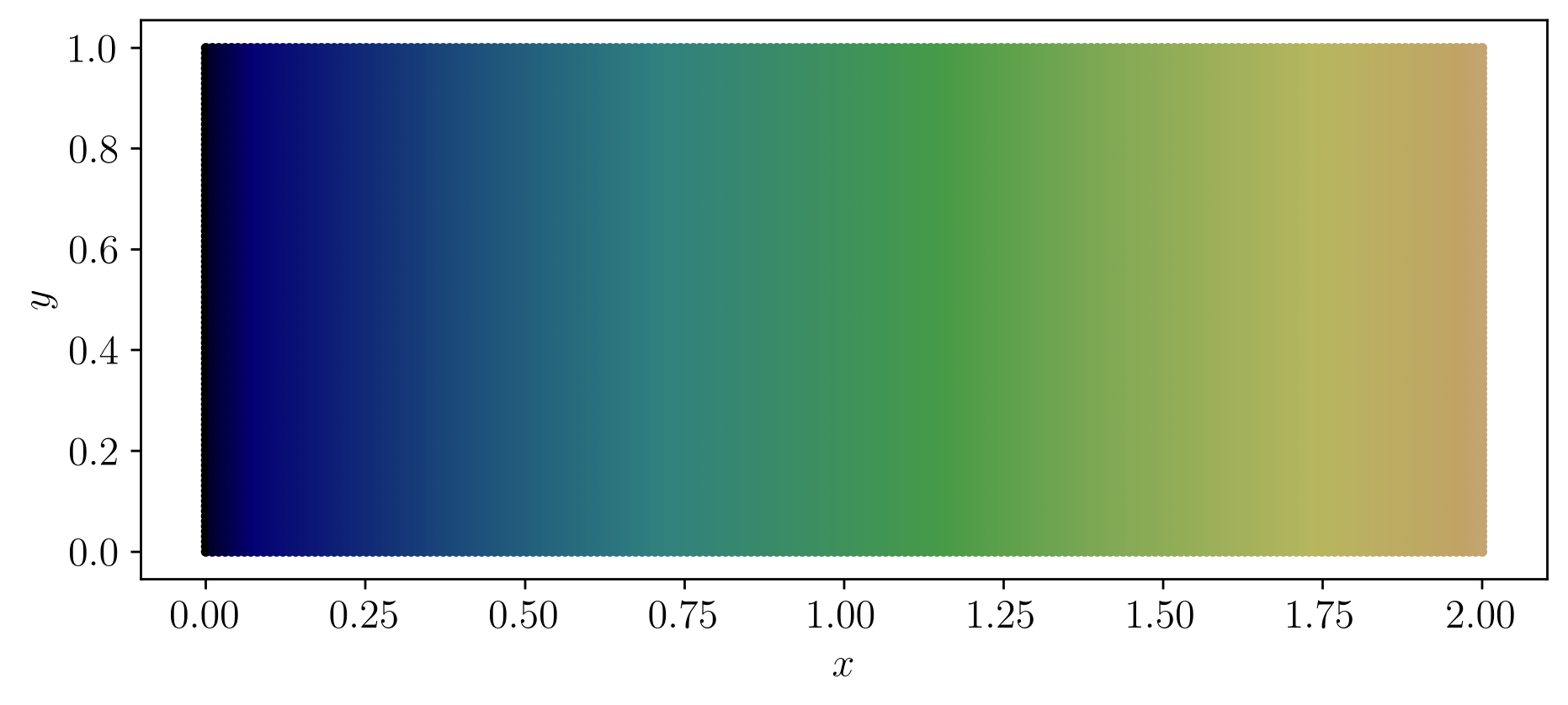}\label{fig:double-gyre-init}} \subfigure{\includegraphics[width=0.32\textwidth]{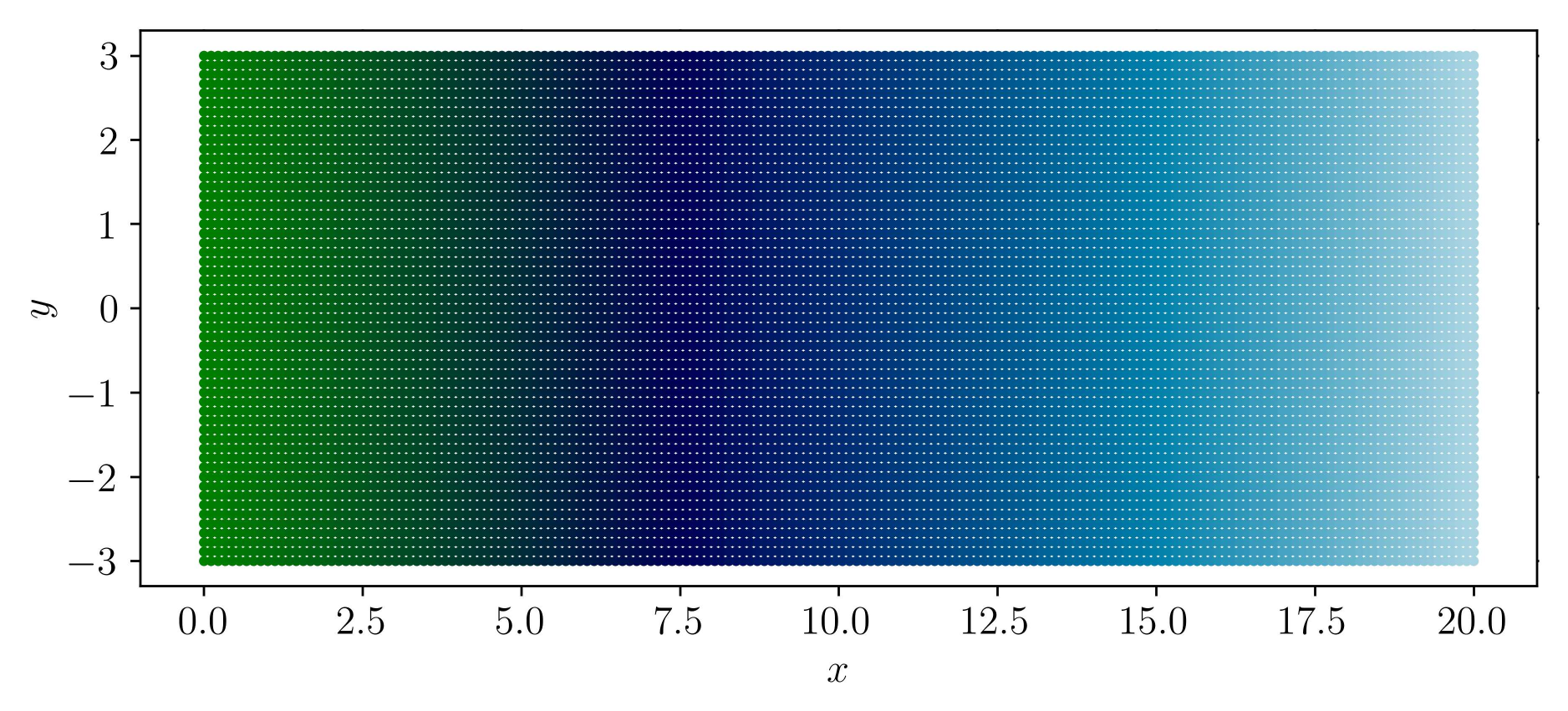}\label{fig:bickley-jet-init}}
    
    \subfigure{\includegraphics[width=0.32\textwidth]{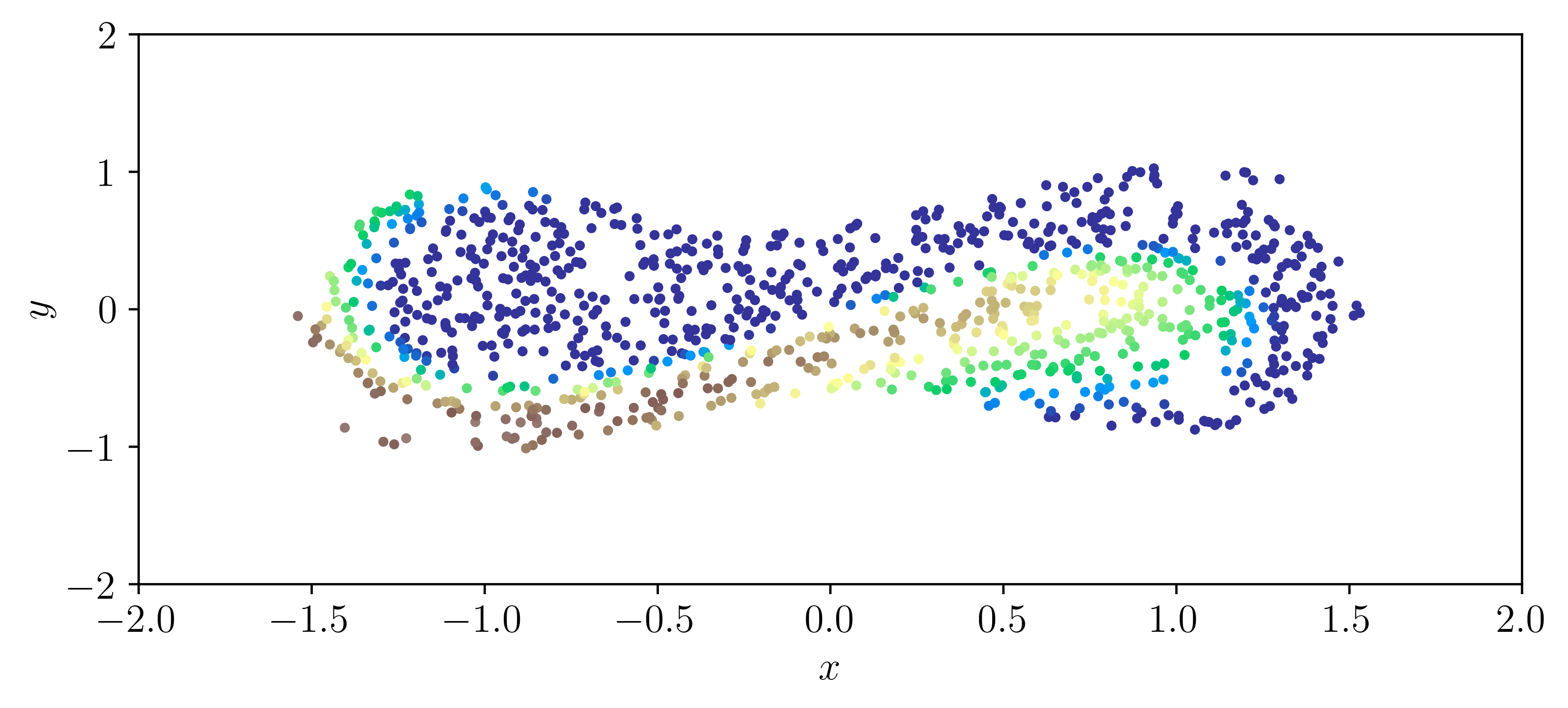}\label{fig:duffing-oscillator-overview}}
    \subfigure{\includegraphics[width=0.32\textwidth]{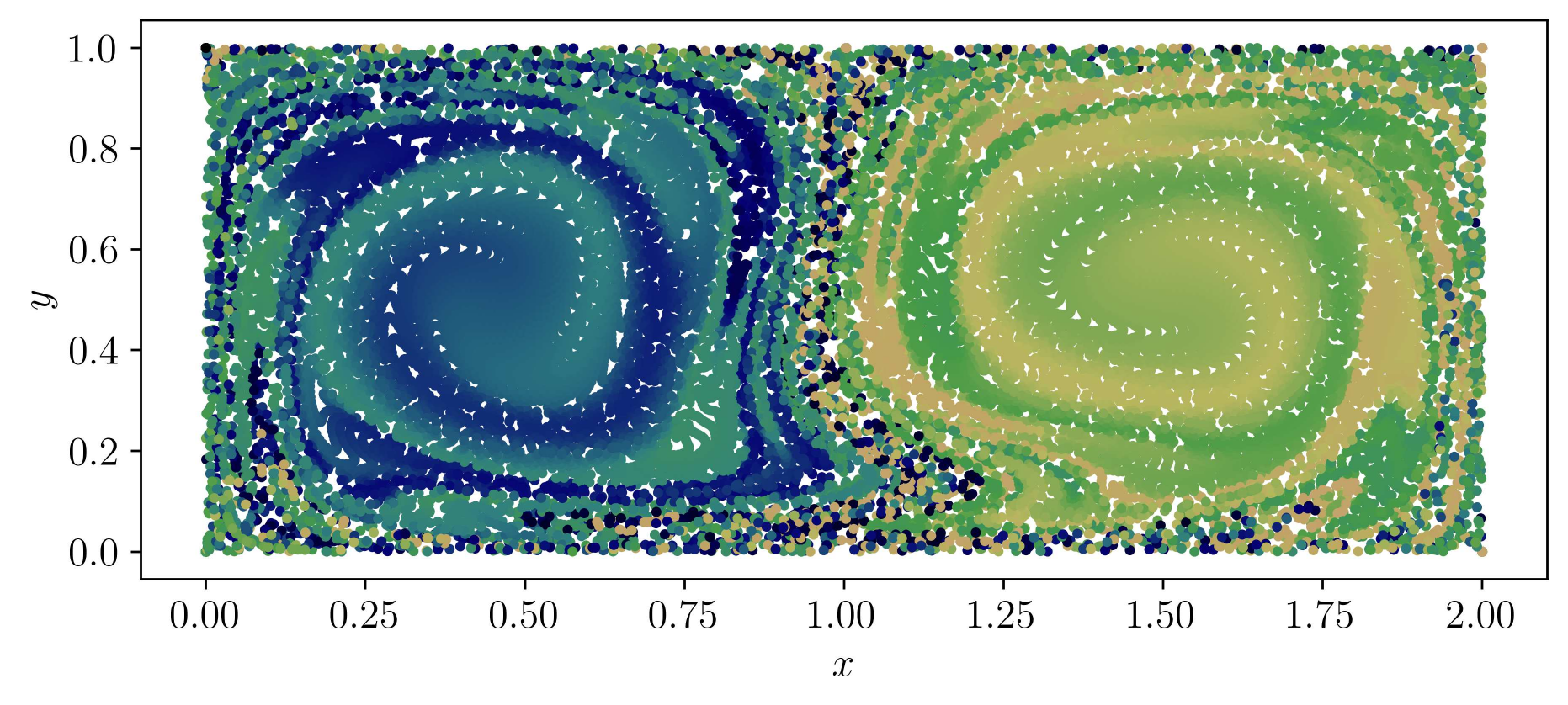}\label{fig:double-gyre-overview}}
     \subfigure{\includegraphics[width=0.32\textwidth]{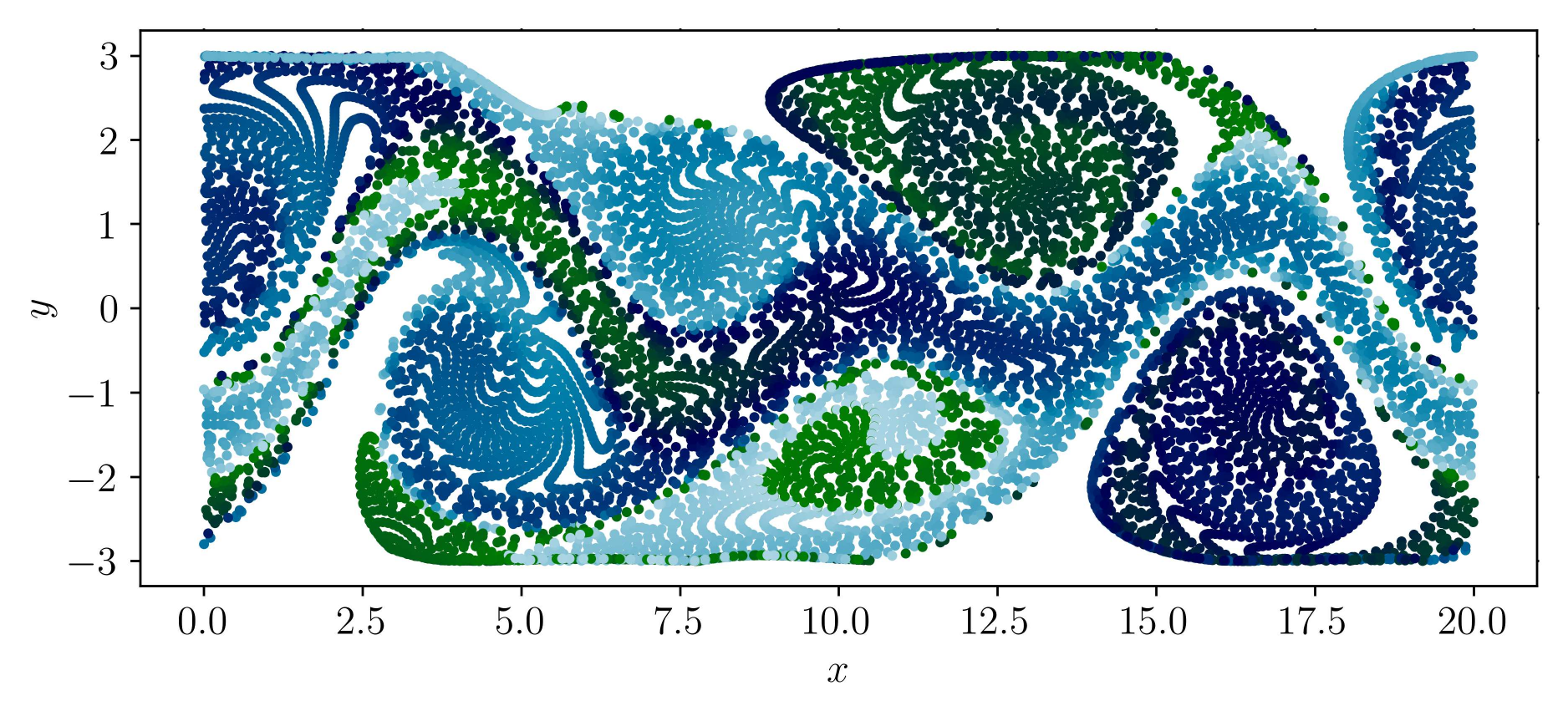}\label{fig:bickley-jet-overview}}
    
    \caption{Prototypical dynamical systems for studying transfer operators. (\textit{a}) $1000$ particles, colored by their initial $x$-coordinate to visualize movement, shown in a $[-2, 2] \times [-2, 2]$ grid. (\textit{d}) Positions of particles are shown at time $t = 2.5$ under the Duffing oscillator, which describes the dynamics of a point mass in a double well potential. (\textit{b}) $20000$ particles are initialized in a $[0, 2] \times [0, 1]$ grid. (\textit{e}) Particles' positions are tracked in a time-dependent double gyre flow at time $t = 12.1$. Two dynamically distinct regions form in this flow, a representative feature of the system. (\textit{c}) $9900$ particles are initialized in a $[0, 20] \times [-3, 3]$ grid. (\textit{e}) Movement of the particles at $t = 10.1$ under a Bickley jet model, an idealized model for fluid flows.}
\end{figure}
\subsection{Simulated Dynamical Systems}
In this subsection, we describe three prototypical dynamical systems used in the study of transfer operators, the Duffing oscillator, the double gyre, and Bickley jet. In the subsequent subsections, we demonstrate the use of the proposed algorithms for the kernel Koopman operator in Section \ref{sec:methodology}.

\subsubsection{Duffing oscillator}
The Duffing oscillator is a nonlinear second order differential equation used to model damped and driven oscillators. The evolution of the particles whose positions are described by $(x, y)$ is governed by the $\dot{x} = y$ and $ \dot{y} = -\delta y - x (\beta + \alpha x^2)$. We use the parameters $\delta = 0.5$, $\beta=-1$, and $\alpha=1$. We simulate $1000$ points uniformly sampled on grid $[-2, 2] \times [0, 1]$, shown in Fig. \ref{fig:duffing-oscillator-init}. From time $t \in [0, 2.75]$ with step size $0.25$, we use a differential equations solver based on an explicit Runge-Kutta (4,5) formula on the velocities to solve an initial value problem for the system of ordinary differential equations with the sampled points. The visualization of the movement of particles in a Duffing oscillator is shown in Fig. \ref{fig:duffing-oscillator-overview}.

\subsubsection{Double gyre model}
A simple model of the wind-driven, time-dependent double gyre flow is described by $f(x, t) = \epsilon * \sin{(\omega t)} * x^2 + (1 - 2 \epsilon \sin{(\omega t})) * x$, $\frac{\partial f}{\partial x} = 2 \alpha * \sin{(\omega t)} * x + (1 - 2 \alpha \sin{(\omega t}))$, $\dot{x} = -\pi A \sin{(\pi f(x,t))} \cos{(\pi y)}$, $\dot{y} = \pi A \cos{(\pi f(x,t))} \sin{(\pi y)} * \frac{\partial f}{\partial x}$.

The parameters are set to $\epsilon = 0.25, \alpha=0.25, A = 0.25, \text{and } \omega = 2 \pi$ \citep{Forgoston2011Set-basedInvariant}. We simulate $20000$ points, as in Fig. \ref{fig:double-gyre-init}, uniformly sampled on grid $[0, 2] \times [0, 1]$. From time $t \in [0, 20]$ with step size $0.1$, we use a differential equations solver based on an explicit Runge-Kutta (4,5) formula on the velocities to calculate the trajectories of the points, shown in Fig. \ref{fig:double-gyre-overview}.

\subsubsection{Bickley jet}
The Bickley jet model is a prototypical model in the study of coherence that is a meandering zonal jet, flanked both above and below by counter rotating vertices. The Bickley jet is used as an idealized model for the Gulf Stream in the ocean and polar night jets in the atmosphere \citep{Del-Castillo-Negrete1992ChaoticFlow, Beron-Vera2010Invariant-tori-likeFlows}. The stream function for the Bickley jet model is 
$\psi(x,y,t) = \psi_0(y) + \psi_1(x,y,t)$,  $\psi_0(y) = -U_0 L_0 \tanh\left(\frac{y}{L_0}\right)$, $\psi_1(x, y, t) = U_0 L_0 \sech^2\left(\frac{y}{L_0}\right) \Re \left({\sum_{n=1}^3 f_n(t) \exp(ik_nx)}\right)$, with $f_n(t) = \epsilon_n \exp(-ik_nc_nt)$. The velocities can be computed as $\dot{x} = {\partial{\psi}}/{\partial{x}}$ and $\dot{y} = {\partial{\psi}}/{\partial{y}}$. We use scaled parameters $U_0 = 5.4138, L_0 = 1.77, c_1 = 0.1446U_0, c_2 = 0.2053U_0, c_3 = 0.4561U_0, \epsilon_1 = 0.075, \epsilon_2 = 0.4, \epsilon_3 = 0.3, r_0 = 6.371, k_1 = 2/r_0, k_2 = 4/r_0, k_3 = 6/r_0$ \citep{Hadjighasem2017ADetection}.  We sample $9900$ points uniformly on a grid $[0, 20] \times [-3, 3]$, seen in Fig. \ref{fig:bickley-jet-init}. From time $t \in [0, 40]$ with step size $0.1$, we calculate the trajectories of the points using a variable-step, variable-order Adams-Bashforth-Moulton solver of orders 1 to 13 on the velocity $[\dot{x}, \dot{y}]$, as shown in Fig. \ref{fig:bickley-jet-overview}. 

\subsection{Evaluation Metrics}
In this section, we discuss two evaluation metrics used to compare the kernel Koopman operator with RFFs against EDMD algorithms with dictionaries selected based on the known properties of the simulated environments. We first evaluate the reconstruction error associated with the known trajectory $\mathbf{x}_i$ of a particle $i$ over multiple time steps. We reconstruct trajectories of the system using the Koopman mode decomposition formula. This is described in the presentation of the EDMD algorithm and other works related to dictionary learning \citep{Williams2015ADecomposition, Li2017ExtendedOperator}. We summarize the procedure here for completeness.

First, we solve for a matrix $\mathbf{B} \in \mathbb{R}^{M \times N}$ such that $\mathbf{x} = (\Phi_{\mathbf{X}} \mathbf{B})^\top$. Let $\mathbf{V}$ be the matrix containing the set of right eigenvectors, $\mathbf{W}^*$ be the matrix contained the set of left eigenvectors associated with $\mathcal{K}$, and $\boldsymbol{\mu}$ be the vector of eigenvalues. Each of the $i$ left eigenvectors of should be scaled such that $\mathbf{w}_i^*\mathbf{v}_i = 1$. The approximate eigenfunction corresponding to the eigenvalue $\mu_j$ of $\mathcal{K}$ is then $\psi_j = \mathbf{v}_{j}^\top\Phi$.  Then, the full state can be reconstructed as $\hat{\mathbf{x}} = (\mathbf{W}^*\mathbf{B})^\top (\Phi_{\mathbf{X}} \mathbf{V})^\top$. To estimate the evolution of the trajectory after a time $t$ has elapsed from the positions of the particles at $\mathbf{x}(0)$, we leverage the linear representation of the Koopman decomposition to construct the trajectory as
\begin{equation} \label{eq:reconstruction}
    \hat{\mathbf{x}}(t) = \boldsymbol{\mu}^t (\mathbf{W}^*\mathbf{B})^\top (\Phi_{\mathbf{X}_0} \mathbf{V})^\top.
\end{equation} 
Finally, we compute the error between predicted trajectories from Eq. \eqref{eq:reconstruction} and true trajectories as
\begin{equation} \label{eq:reconstruction-error}
    e_{p} = \sqrt{\frac{1}{T}\sum_{t=1}^T \norm{\mathbf{x}(t) - \hat{\mathbf{x}}(t)}_F^2}.
\end{equation}

\begin{table}
\small
\centering
\caption{Comparison of Error Reconstruction with Different Dictionaries for Various Environments}
\begin{tabular}{|cc|cc||cc||cc|}
\hline
\multicolumn{2}{|c|}{System}                & \multicolumn{2}{c||}{Duffing Oscillator}      & \multicolumn{2}{c||}{Double Gyre}      & \multicolumn{2}{c|}{Bickley Jet}      \\ \hhline{|========|}
\multicolumn{2}{|l|}{Prediction Horizon}    & \multicolumn{1}{c|}{NT} & LT & \multicolumn{1}{c|}{NT} & LT & \multicolumn{1}{c|}{NT} & LT \\ \hhline{|========|}
\multicolumn{1}{|l|}{\multirow{3}{*}{\STAB{\rotatebox[origin=c]{90}{Dict.}}}} & Learned & \multicolumn{1}{l|}{0.5929}   &  0.3713  & \multicolumn{1}{l|}{0.2307}   &  0.7125  & \multicolumn{1}{l|}{2.7584}   &  13.856 \\ \cline{2-8} 
\multicolumn{1}{|l|}{}                  & Gaussian & \multicolumn{1}{l|}{0.4668}   &   0.3427 & \multicolumn{1}{l|}{0.3610}   &   1.0896 & \multicolumn{1}{l|}{9.2036}   &  20.441  \\ \cline{2-8} 
\multicolumn{1}{|l|}{}                  & Monomial & \multicolumn{1}{l|}{0.5466}   &  0.3523  & \multicolumn{1}{l|}{0.3277}   &  0.8891   & \multicolumn{1}{l|}{5.2428}   &   13.524  \\ \hline
\end{tabular} \label{table:error-recons}
\end{table} 

\begin{figure}
    \centering
    \subfigure{\includegraphics[width=0.98\textwidth]{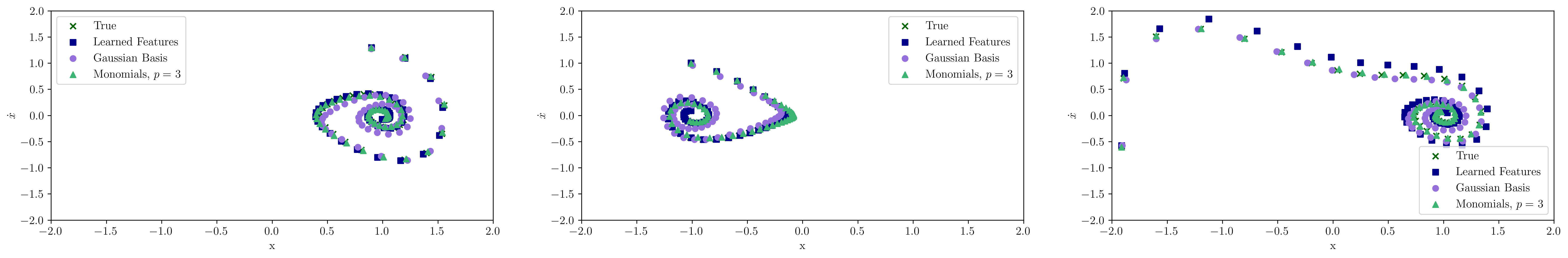}} 
    
    \subfigure{\includegraphics[width=0.98\textwidth]{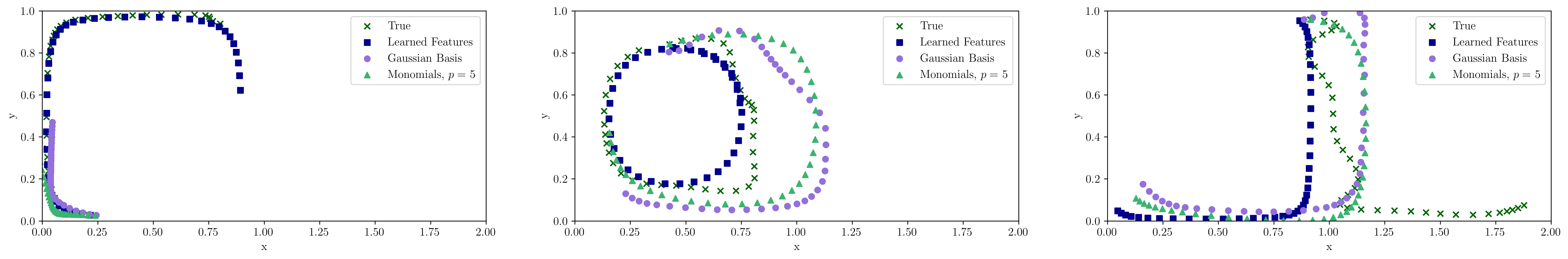}}

    \subfigure{\includegraphics[width=0.98\textwidth]{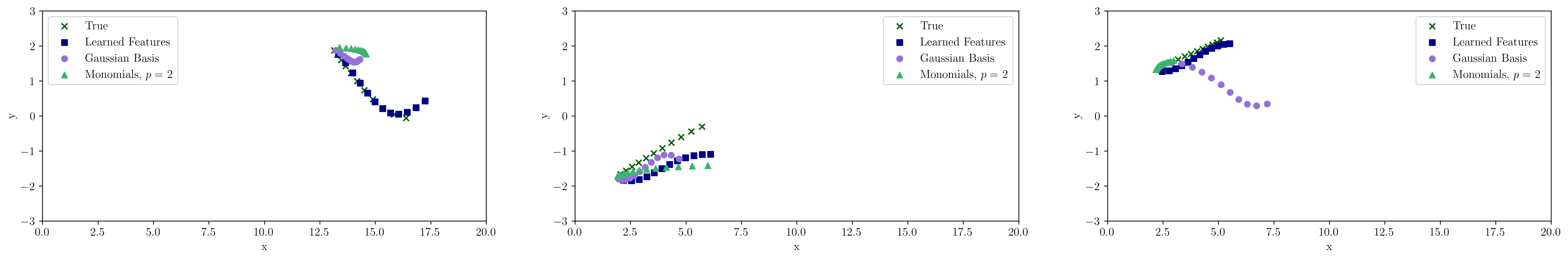}} 

    \caption{Reconstruction of trajectories in the (a) Duffing oscillator, (b) double gyre, and (c) Bickley jet using the Koopman operator constructed from learned features and known dictionaries.}
    \label{fig:reconstruction-dg}
\end{figure}
\begin{figure}[!ht]
    \centering
    \subfigure{\includegraphics[width=0.32\textwidth]{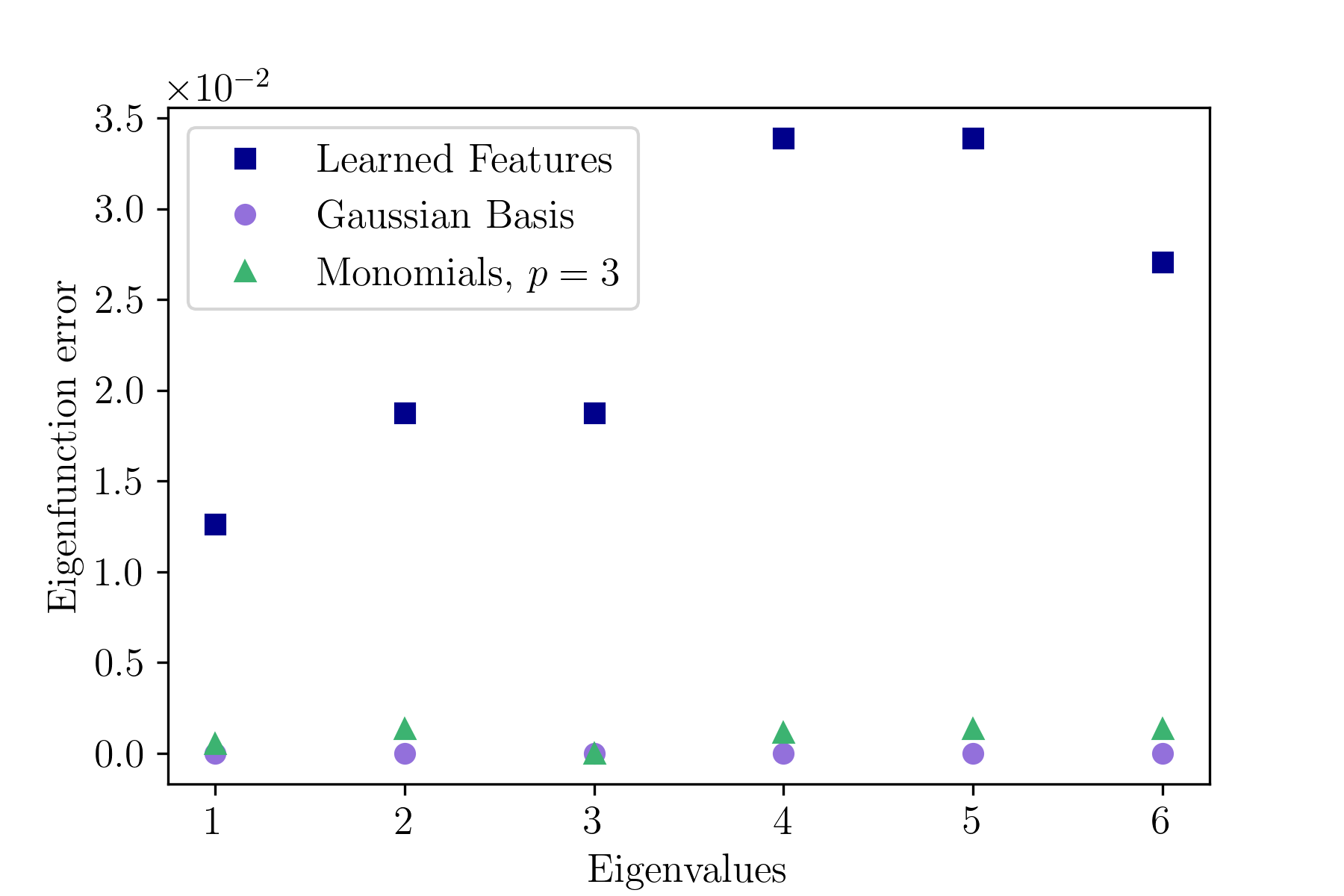}} \subfigure{\includegraphics[width=0.32\textwidth]{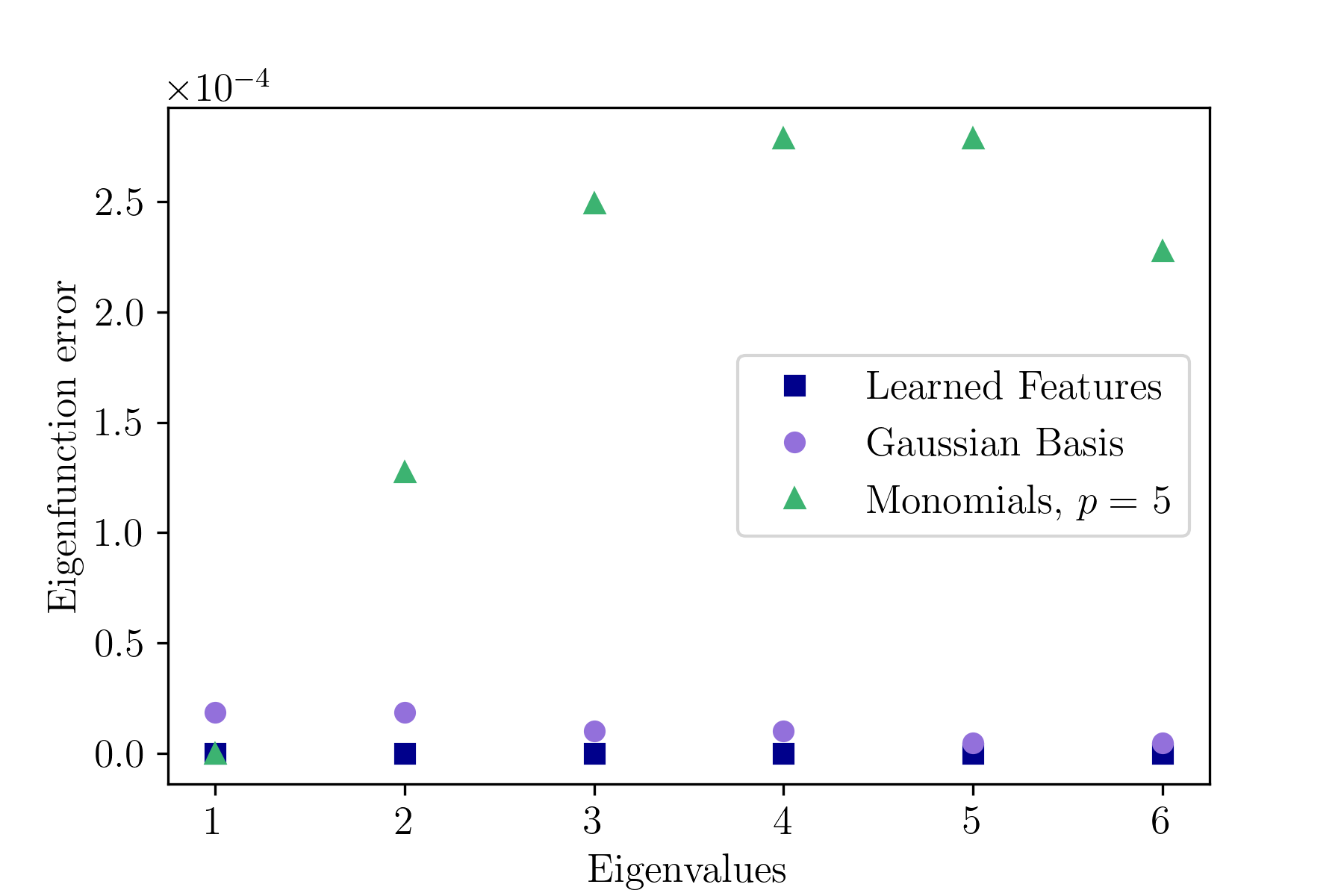}}
    \subfigure{\includegraphics[width=0.32\textwidth]{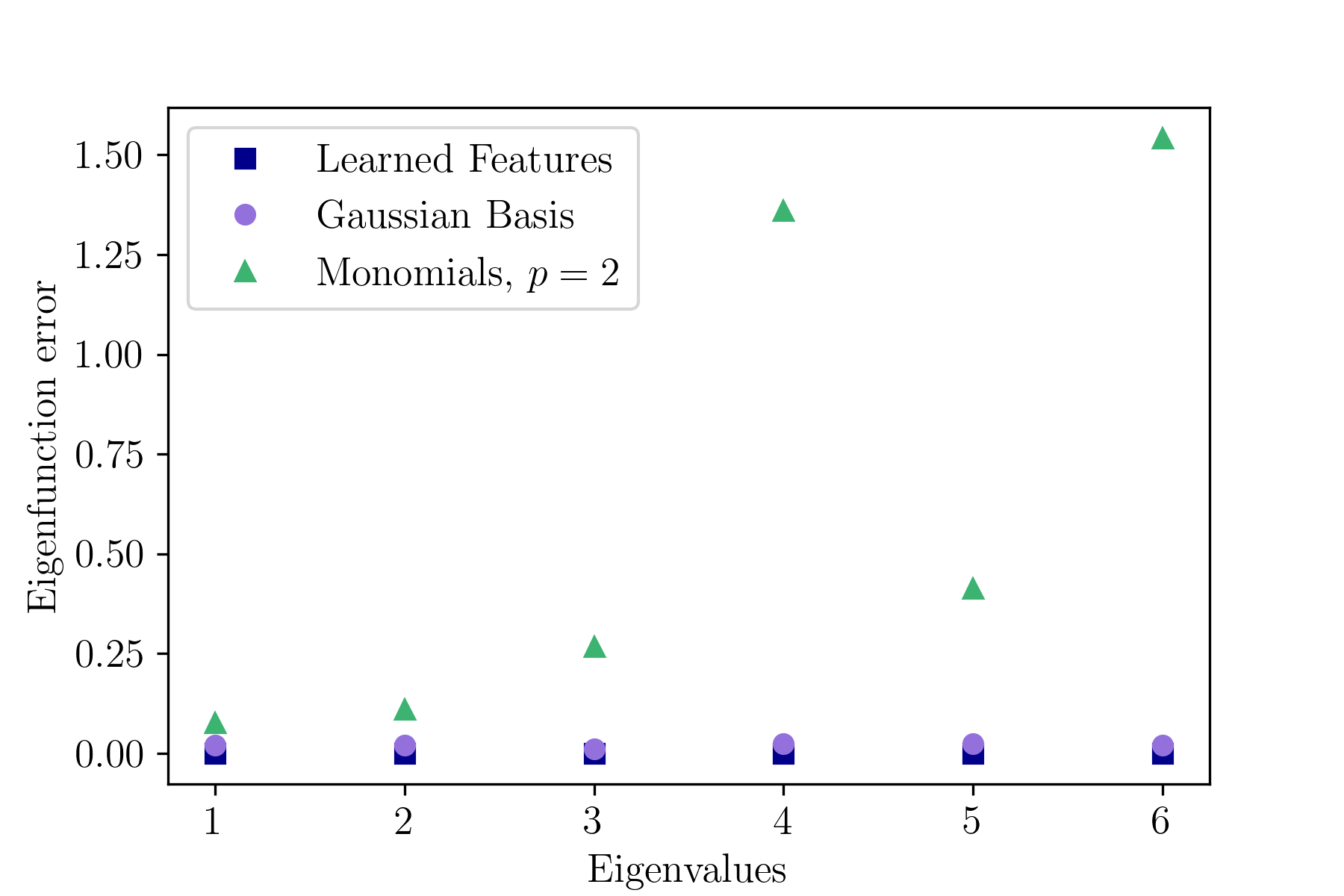}}
    \caption{Eigenfunction error approximation for Koopman operator constructed with learned features and known dictionaries. Errors were computing using randomly samples points in the (a) Duffing oscillator, (b) double gyre, and (c) Bickley jet environments and averaging their corresponding eigenfunction error.} \label{fig:eigenvalues-error}
\end{figure}

For all examples, we analyze near-term and long-term prediction errors in the reconstruction. For the learned features, we set up the optimization framework using Eq. \eqref{eq:min-loss}. To compare against our framework, we construct the approximation of the Koopman operator through the EDMD algorithm \citep{Williams2015ADecomposition}. Specifically, we use the insights from previously studied EDMD algorithms on the Duffing oscillator, double gyre, and Bickley jet to construct dictionaries \citep{Li2017ExtendedOperator, Kaiser2021Data-drivenControl, Salam2022LearningAwareness}. For the Duffing oscillator, we compare our 100 RFFs against dictionaries created using the Gaussian basis function (GBF) on a $50 \times 50$ discretization of the system with $\sigma=1e^-4$ and monomial basis functions of degree $3$. For the double gyre, we use 100 RFFs, and the baseline has GBFs on a $10 \times 5$ discretization with $\sigma=0.1$ and monomial basis functions of degree $5$. For the Bickley jet, we use 200 RFFs, while the dictionaries are GBFs with $\sigma=1.1$ on a $10 \times 10$ discretization and monomial basis function of degree $2$. While the learned features are computed online over the entire available dataset, the EDMD algorithm is computed against each snapshot pair of data $\mathbf{X}(t)$ and $\mathbf{X}(t+1)$. The estimated near-term (NT) trajectory is found using Eq. \eqref{eq:reconstruction} for $10$ time steps into the future, as in we reconstruct $[\hat{\mathbf{X}}(t+2), \dots \hat{\mathbf{X}}(t+11)].$ For the long-term (LT) estimation, we compute $40$ time steps into the future. Finally, the error is computing using the Eq. \eqref{eq:reconstruction-error} for trajectory prediction errors. 

The results of the error reconstructions are shown in Table \ref{table:error-recons}. The errors shown in the table represent the difference between the true trajectory and trajectories from the Koopman operator constructed from dictionaries with the learned features, the Gaussian basis functions, and the monomial basis functions. We see for simpler systems, such as the Duffing oscillator, the learned features perform comparably, if not slightly worse, than models from handcrafted dictionaries. However, for more complicated systems, such as the double gyre and Bickley jet, reconstruction in the long-term regime using the learned features outperforms the methods using the handcrafted dictionaries. This is exemplified most clearly in Fig. \ref{fig:reconstruction-dg} for trajectories of randomly sampled particles. For short-term predictions, the trajectories of the particles estimated using the learned features versus the Gaussian or monomial basis functions are similar. However, as more time elapses between the initial position of the particle and the estimate, the learned features are able to more closely match the true trajectory. The trajectories estimated using the Gaussian basis functions diverge after several time steps. While other basis functions capture some notion of coherence, the learned features are able to capture both the notion of coherence and the movement of individual particles within the space. 

We next evaluate the  accuracy of the eigenfunction approximation \citep{Li2017ExtendedOperator}. Let the approximate eigenfunction corresponding to the eigenvalue $\mu_j$ of $\mathcal{K}$ be $\psi_j = \mathbf{v}_{j}^\top\Phi$. Then, the accuracy of the eigenfunction approximation is
\begin{equation} \label{eq:eigen-error}
    e_{f_j} = \sqrt{\frac{1}{I}\sum_{i=1}^I |\psi_j(f(x(i)) - \mu_j \psi_j(x(i))|^2}.
\end{equation}

We show the eigenfunction approximation for the Duffing oscillator and double gyre comparisons for various features and dictionaries in Fig. \ref{fig:eigenvalues-error}. The value of $e_{j}$ as in Eq. \eqref{eq:eigen-error} are selected for the first few eigenvalues in all examples. We select $100$ randomly sampled points for each dynamical system and calculate the eigenfunction approximation error for the eigenfunctions from the learned model, the Gaussian basis functions, and the monomial basis functions. We can see the learned features perform comparably to the monomial basis functions, which also provided good reconstruction. The learned features also outperform the Gaussian basis functions. Note, that the low error in approximation is an indicator of the invariance of the eigenfunctions, as it shows that the application of eigenfunctions do not cause high divergence.

\subsection{Quantitative features}
\begin{figure}
    \centering  \subfigure{\includegraphics[width=0.23\textwidth]{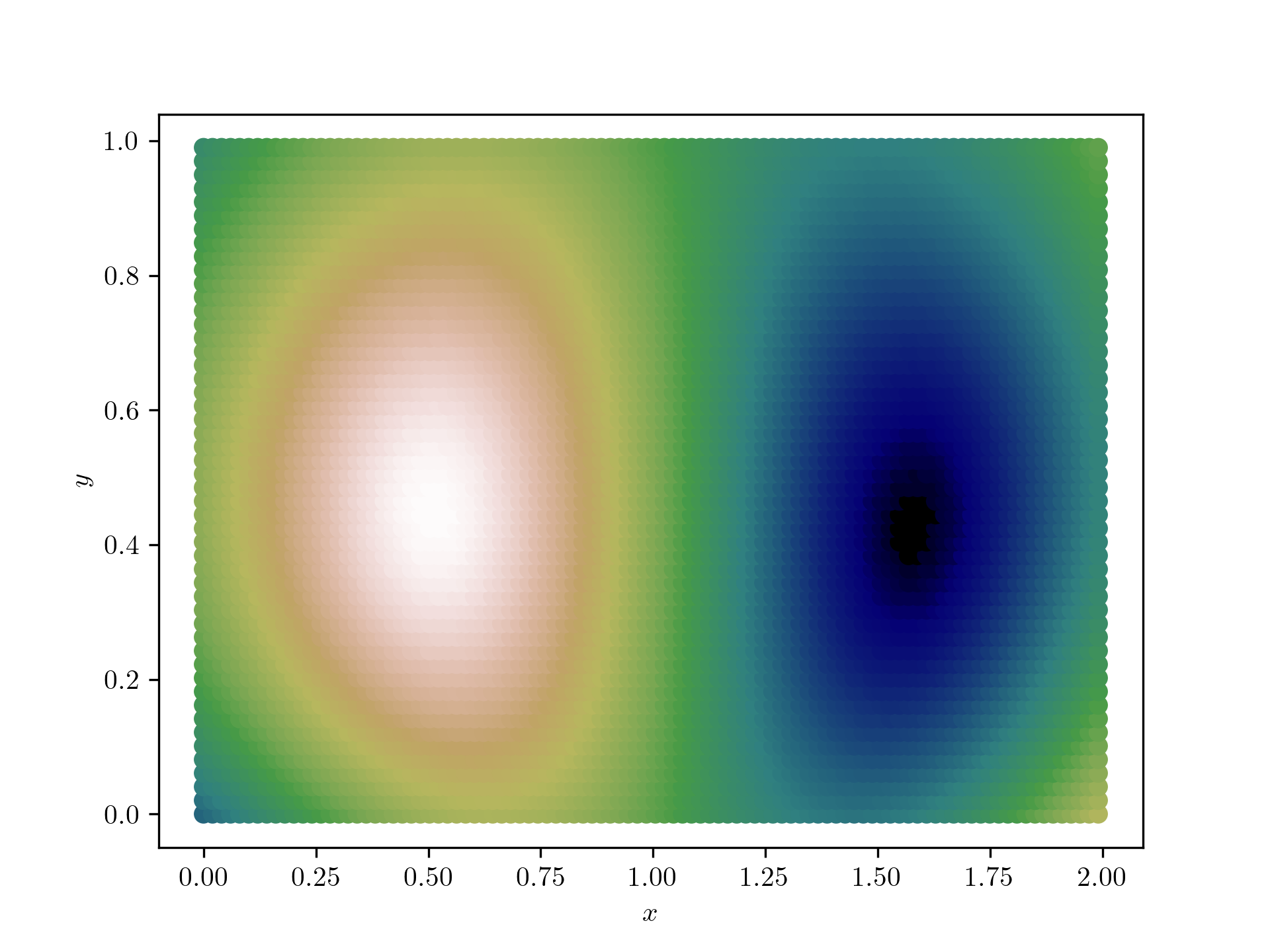}\label{fig:kernel-efunc-dg}}
    \subfigure{\includegraphics[width=0.23\textwidth]{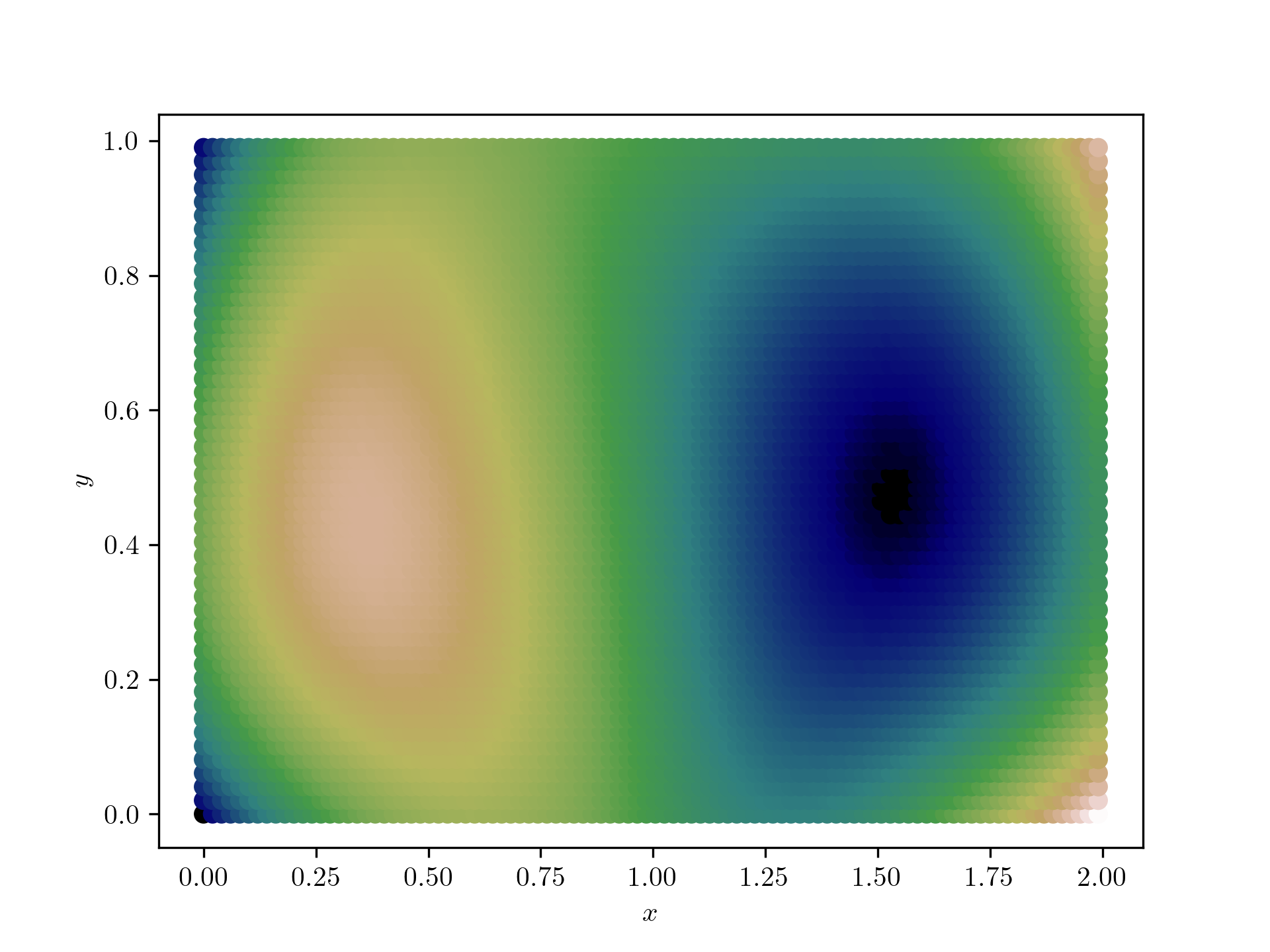}\label{fig:learned-efunc-dg}}
    \subfigure{\includegraphics[width=0.21\textwidth]{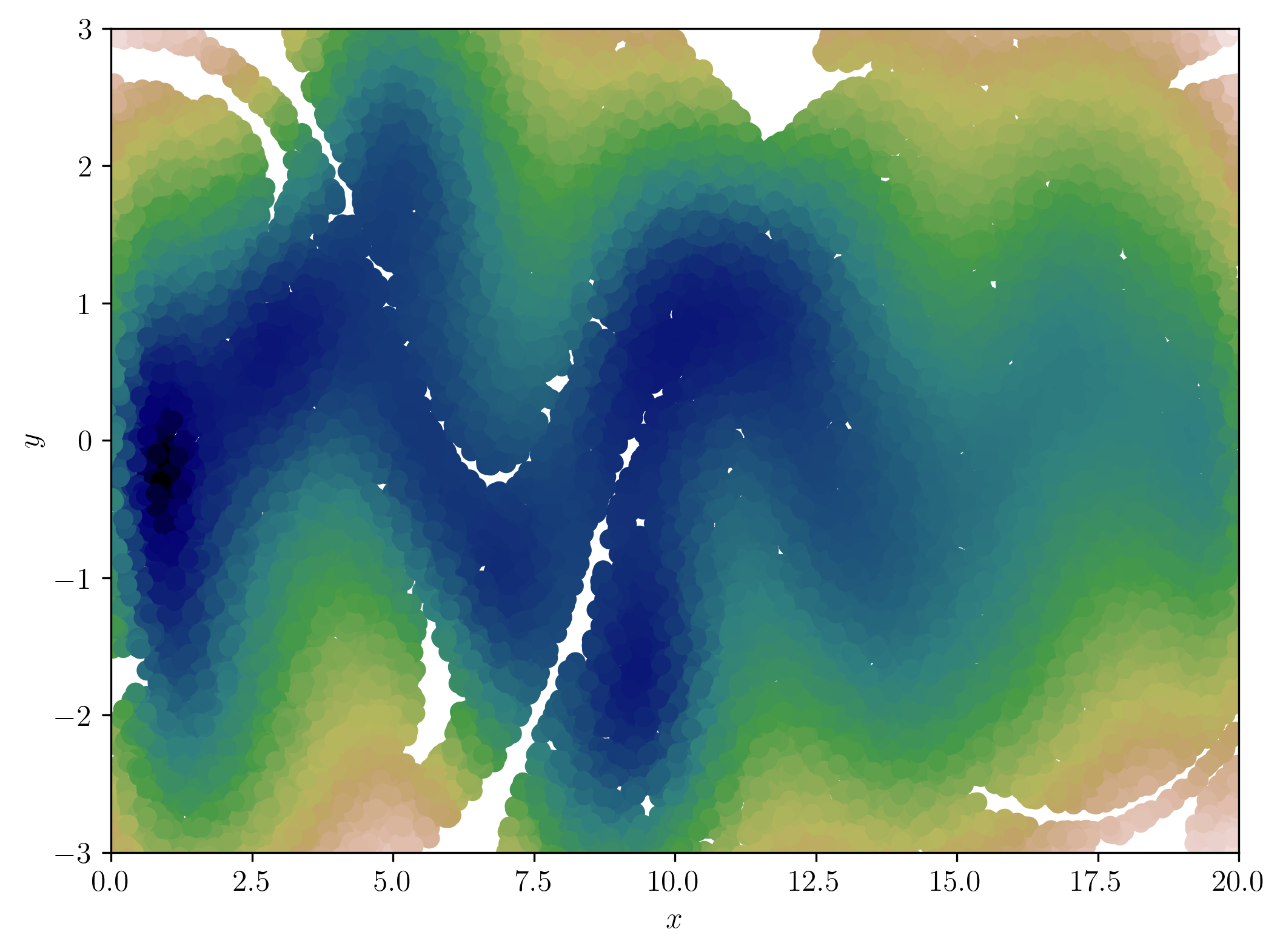}\label{fig:kernel-efunc-bj}}
    \subfigure{\includegraphics[width=0.23\textwidth]{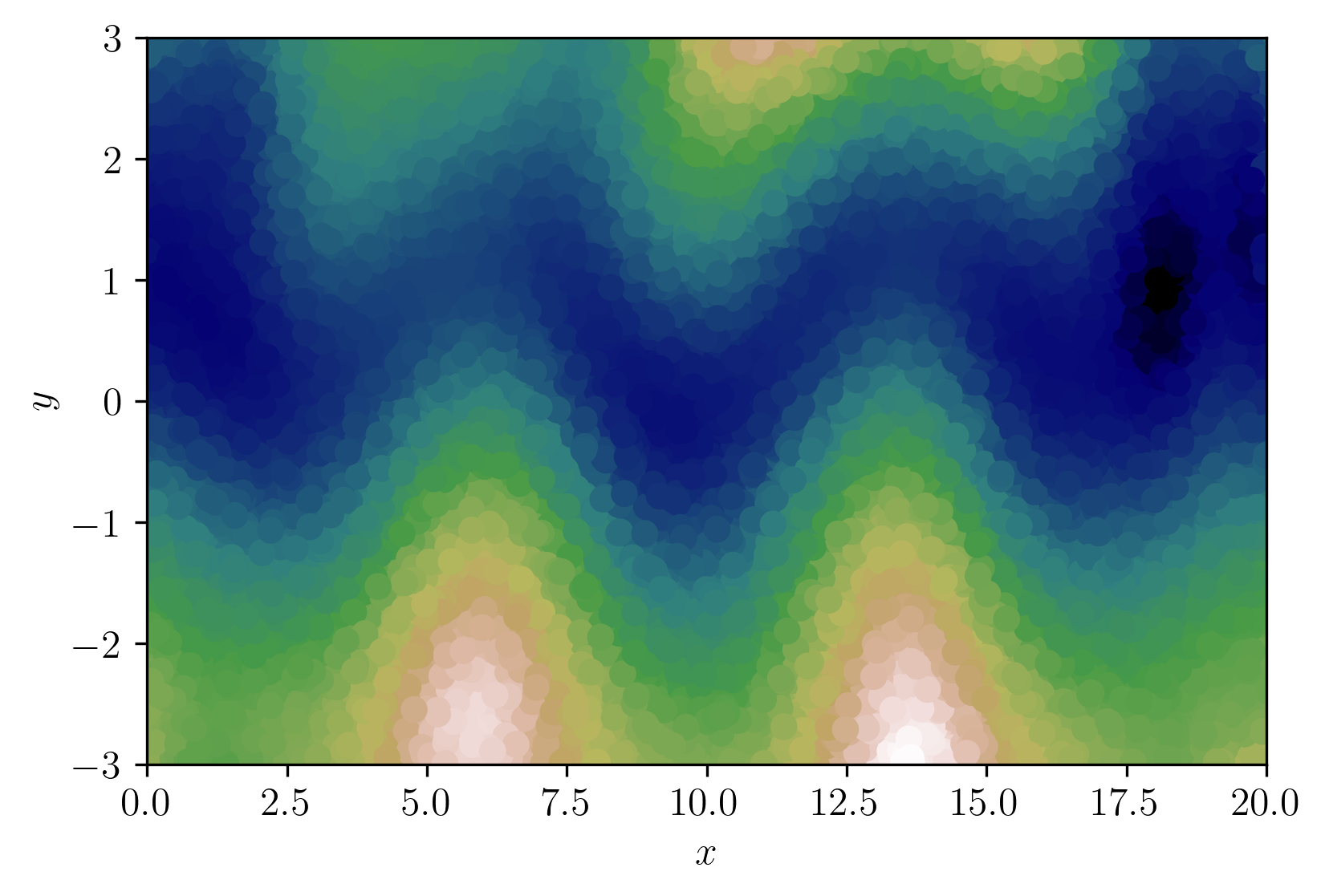}\label{fig:learned-efunc-bj}}

    \caption{Eigenfunction comparison for kernel EDMD algorithm versus EDMD with learned features for time-dependent double gyre and Bickley jet. Dominant eigenfunction (\textit{a}) from the Koopman operator constructed using kernel EDMD and (\textit{b}) using learned features for double gyre. Dominant eigenfunction (\textit{c}) from the Koopman operator constructed using kernel EDMD and (\textit{d}) using learned features for Bickley jet. } \label{fig:eigenfunc-viz}
\end{figure}

Next, we quantitatively compare eigenfunction values derived from the EDMD using the learned RFFs with the kernel EDMD algorithm \citep{Williams2015AAnalysis}. For the double gyre and Bickley jet, we know that the coherent sets, an important property of global dynamics, can be estimated using an RBF kernel with $\sigma = 0.75$ and $\sigma = 1.0$, respectively \citep{Salam2022LearningAwareness}. We plot the eigenfunctions of the Koopman operator associated with the dominant eigenvalues of the decomposition. The eigenfunctions of the Koopman operator corresponding to the kernel EDMD formulation elucidate two distinct regions for the double gyre and waves for the Bickley jet, as seen in Fig. \ref{fig:kernel-efunc-dg}  and \ref{fig:kernel-efunc-bj} respectively. The eigenfunctions of the Koopman operator learned using the proposed algorithm in Fig. \ref{fig:learned-efunc-dg} and \ref{fig:learned-efunc-bj} agree closely with verified notions of coherence within these systems.

%% file: sections/conclusion.tex
In this paper, we develop an automated, online procedure for constructing operators to describe global representations of dynamical systems. This work can be generalized through the use of non-stationary kernels and extended to applications in controls. Non-stationary kernels capture input-dependent correlations. While non-stationary kernel learning gained attention in recent years \citep{Samo2015GeneralizedKernels, Remes2017Non-stationaryKernels}, the problem of automated non-stationary kernel is still an active area of research. While transfer operators are well-studied in controls, there are practical considerations in applying these learned models to control theory. The dictionary of observables may fail to satisfy the invariance property for observables and the derived evolution of the observable may leave the lifted space \citep{Bruder2019ModelingControl}. While several works focus on finding functions that are the basis of Koopman invariant subspaces \citep{Kaiser2021Data-drivenControl, Kawahara2016DynamicAnalysis}, there are no mathematical guarantees for the identified functions to be Koopman eigenfunctions \citep{Haseli2021LearningDecomposition}. Notably, many open questions surround the representation and evolution of control inputs in transfer operator methods \citep{Proctor2018GeneralizingControl, Otto2021KoopmanSystems}.

%% file: sections/acknowledgements.tex
We greatly acknowledge the support of the NSF IUCRC 1939132, NSF IIS 1910308, and the University of Pennsylvania's University Research Foundation Award.